\documentclass{article}

\usepackage{arxiv}

\usepackage{amsmath,amssymb}

\usepackage[utf8]{inputenc} 
\usepackage[T1]{fontenc}    
\usepackage{hyperref}       
\usepackage{url}            
\usepackage{booktabs}       
\usepackage{amsfonts}       
\usepackage{nicefrac}       
\usepackage{microtype}      
\usepackage{lipsum}		
\usepackage{graphicx}
\usepackage{natbib}
\usepackage{doi}

\hypersetup{
  pdftitle={Spatial machine-learning model diagnostics: a model-agnostic distance-based approach},
  pdfauthor={Alexander Brenning},
  pdfkeywords={interpretable machine learning, spatial prediction error, spatial variable importance, spatial cross-validation},
  hidelinks}
\usepackage{graphicx}

\begin{document}

\title{Spatial machine-learning model diagnostics: \\
 a model-agnostic distance-based approach}

\author{Alexander Brenning}

\date{Friedrich Schiller University Jena, Department of Geography and Michael Stifel Center Jena for Data-Driven and Simulation Science (MSCJ), Jena, Germany}

\maketitle

\begin{abstract}
While significant progress has been made towards explaining
black-box machine-learning (ML) models,
there is still a distinct lack of diagnostic tools that elucidate the spatial
behaviour of ML models in terms of predictive skill and variable importance.
This contribution proposes spatial prediction error profiles (SPEPs) and
spatial variable importance profiles (SVIPs) as novel model-agnostic
assessment and interpretation tools for spatial prediction models
with a focus on prediction distance.
Their suitability is demonstrated in two case studies representing a
regionalization task in an environmental-science context, and a classification
task from remotely-sensed land cover classification. In these case studies, the
SPEPs and SVIPs of geostatistical methods, linear models, random forest,
and hybrid algorithms show striking differences but also relevant similarities.
Limitations of related cross-validation techniques
are outlined, and the case is made that modelers should focus their model
assessment and interpretation on the intended spatial prediction horizon.
The range of autocorrelation, in contrast, is
not a suitable criterion for defining spatial cross-validation test sets.
The novel diagnostic tools
enrich the toolkit of spatial data science, and may improve ML model
interpretation, selection, and design.
\end{abstract}


\renewcommand{\shorttitle}{Spatial machine-learning model diagnostics}

\section{Introduction}\label{introduction}

Machine-learning (ML) and hybrid geostatistical--ML models such as regression--kriging have become increasingly popular in spatial prediction modeling \citep[for example,][]{hengl.et.al.2015.rfrk, sekulic.et.al.2020}.
While parametric geostatistical techniques such as kriging provide estimates of predictive uncertainty that are backed by statistical theory, most ML models do not, and therefore computational estimation procedures are needed, which must be adapted to the spatial context \citep{brenning.2012.sperrorest, rest2014, roberts2017}.
Similarly, the interpretation of complex black-box ML models can be challenging, and explicitly spatial perspectives are still limited.
Overall, we can therefore detect a remarkable lack of model-agnostic computational diagnostics for an explicitly spatial model assessment and interpretation.
The present contribution aims to enrich this field by offering novel tools and perspectives.

In \emph{spatial model assessment}, (non-spatial) leave-one-out cross-validation (LOO-CV) and cross-validation in general have long been used in the context of geostatistical regionalization \citep{isaaks.srivastava.1989, goovaerts.2000.interpolation, webster.oliver.2007, fouedjio.klump.2019}, but they do not provide a spatially differentiated summary.
Their results can furthermore be misleading when the spatial distribution of observations is uneven \citep{isaaks.srivastava.1989}, as, for example, weather stations tend to be clustered in densely populated regions.
Spatial adaptations proposed so far have partitioned the study region into spatially disjoint training and validation areas \citep{russ.brenning.2010, brenning.2012.sperrorest, bahn.mcgill.2013}.
Some studies have proposed to establish a distance buffer between training and test sets or a spatial block size based on the ---or some sort of--- autocorrelation distance \citep{brenning2005, rest2014, roberts2017, valavi2019}, an \emph{ad hoc} practice that needs to be critically discussed.
Considering these recent developments, there's a clear need and opportunity to establish model-agnostic model assessment tools that detect if and how the predictive performance of different models deteriorates as the prediction distance increases.

In \emph{spatial model interpretation}, permutation-based variable importance (PVI) is a popular tool for interpreting ML models. 
Given its limitations when predictors are strongly dependent \citep{hooker.mentch.2019.stop, molnar.2019.iml.book}, modifications have been proposed to interpret predictor effects in transformed space \citep{brenning.2021.wiml}, or to examine conditional importance measures \citep{strobl.et.al.2008.conditional}.
A spatial adaptation of PVI has been proposed to measure how much a predictor contributes to the model's skill in different locations, such as adjacent regions \citep{russ.brenning.2010.svi}.
Nevertheless, this spatial diagnostic does not yet show how each variable's contribution deteriorates or increases with spatial distance, depending on a model's structure and capabilities.
To better elucidate such effects, this contribution proposes to extend the spatial PVI to a continuous distance scale in order to construct spatial variable importance profiles (SVIPs) as a novel model-agnostic interpretation tool for ML models.

This paper is organized around the two proposed ideas and two case studies. 
Specifically, the following section introduces the fundamental concepts of the proposed spatial model assessment and model interpretation tools.
Two case studies are then introduced, representing a regionalization (or spatial regression) problem from environmental science, and a classification task from remote sensing.
The results are finally discussed regarding their utility, their relationships to other resampling-based as well as theoretically derived performance estimates, and their broader implications.

\section{Proposed method}\label{proposed-method}

\subsection{Spatial leave-one-out for model assessment}\label{spatial-leave-one-out-for-model-assessment}

In spatial prediction of categorical response variables (i.e., classification) and quantitative response variables (i.e., regression or regionalization), we use a model \(\widehat{M}\) to predict unobserved response values based on observed values of \(p\) predictor variables or features, \(\mathbf{f}=(f^{(1)}, \ldots, f^{(p)})^T\).
The model is trained on a training sample \(L\) comprising \(n\) observations of response and predictors, and its performance is estimated on a test sample or with a cross-validation (CV) or bootstrap estimator \citep{efron1983}, including spatial resampling approaches \citep{brenning.2012.sperrorest} and leave-one-out cross-validation (LOO-CV).

In LOO-CV, one observation \(\mathbf{o}_i := (y_i, \mathbf{f}_i)\in L\) at a time is removed from the dataset in order to use it for error estimation, while the remaining data serves as the training sample \(L_{-i}:=L\setminus\{\mathbf{o}_i\}\) for training a model \(\widehat{M}_i\).
Each of these \(n\) models is fitted for the sole purpose of predicting \(y_i\).
This prediction is denoted by \(\hat{y}_i^{(-i)}\) to emphasize the removal of observation \(i\) from the training sample.
The LOO error is calculated by comparing these \(n\) predictions to the observed responses \(y_1, \ldots, y_n\) by means of some error (or accuracy) measure such as the root-mean-square error (RMSE) or the misclassification error rate.
This LOO estimate is referred to as \(\widehat{err}_L^{loo}(M)\).

LOO estimation has been used in the comparison of regionalization models since it is implicitly spatial due to the spatial separation of training and test locations \citep[for example,][]{goovaerts.2000.interpolation}.
Nevertheless, this estimator exerts no direct control on the separation distance, and the mean nearest-neighbour distance may be substantially smaller than the mean prediction distance, especially when observations are spatially clustered \citep{isaaks.srivastava.1989}.

Several authors have proposed to enforce spatial exclusion buffers around the LOO test locations, making reference to the concepts of spatial autocorrelation and statistical independence \citep{brenning2005, rest2014, roberts2017, pohjankukka2017, veronesi.schillaci.2019}; the theoretical shortcomings of such requirements are discussed later in Section \ref{discussionscale}.

From a practical perspective, however, it is critical to know how well a model is able to predict the response at relevant prediction distances that occur in the application of the model.
The range of relevant distances may depend on the size of the gaps between point observations, or the maximum distance of regions to which the model is to be transferred.

For this purpose, spatial LOO with buffer, or simply \emph{spatial LOO} in this study, is formally defined as follows, and without imposing \emph{a priori} limitations on the separation distance, or prediction horizon.

In spatial LOO, the training sample for the \(i\)th iteration is defined as
\[
L_{-D(i,r)} := L\setminus D(i,r),
\]
where \(D(i,r)\) is the subset of \(L\) located within a spatial distance \(\le r\) from the spatial location \(\mathbf{x}_i\) of observation \(\mathbf{o}_i\).
The actual spatial separation distance \[
d_{i,r}:=\mathrm{min}\{|\mathbf{x}_i-\mathbf{x}_j|:\mathbf{o}_j\in L_{-D(i,r)}\}
\]
may be (usually only slightly) greater than the specified \(r\), depending on the spatial distribution of observations.
Therefore, the \(d_{i,r}\) (and not \(r\) itself) are recorded as the prediction distances, along with the predictions \(\hat{y}_i^{(-D(i,r))}\).
The recorded values are denoted by \(\hat{y}_{(k)}\), \(y_{(k)}\) and \(d_{(k)}\), and the spatial LOO error, calculated from values recorded at an approximate distance of \(r\), is referred to as
\[
\widehat{err}_L^{(r)}(M).
\]

Note that for \(r=0\), this spatial LOO becomes equivalent to conventional LOO if (and only if) all observations are at unique locations.
In order to embed LOO within the spatial LOO framework, it is therefore convenient to define
\(D(i,r):=\{\mathbf{o}_i\}\) for \(r<0\).

\subsection{Spatial prediction error profiles (SPEPs)}\label{spatial-prediction-error-profiles-speps}

In order to visualize the average relationship between prediction error and distance, a spatial LOO error \(\widehat{err}_L^{(r)}(M)\) needs to be estimated as a function of prediction distance.
For this purpose, the recorded distances \(d_{(k)}\) are binned.
Within each of these bins, denoted by an index set \(B=\{k_1,\ldots,k_c\}\), the performance \(\widehat{err}_L^{(B)}(M)\) is calculated from all corresponding \(\hat{y}_{(k)}\) and \(y_{(k)}\) values, \(k\in B\).
The lag distance \(\hat{d}_B\) assigned to this estimate is the median of the recorded distances, \(\mathrm{median}\{d_{(k)}:k\in B\}\).

The resulting series of \((\hat{d}, \widehat{err})\) values is referred to as a \emph{spatial prediction error profile} (\emph{SPEP}). It allows us to address the following key questions of spatial prediction modeling:

\begin{itemize}
\item
  How does model performance deteriorate with increasing distance from the training data? In other words, how well does a model fill local gaps, and how well does it extrapolate?
\item
  How do models differ in their ability to predict at small and greater distances?
\end{itemize}

It shall be noted that training sample size may decrease substantially as the separation distance \(r\) increases, which in itself may result in a drop in model performance and in some cases a biased or poorly representative distribution of the remaining data.
Also, individual observations can be used multiple times in the calculation of \(\widehat{err}_L^{B}(M)\) for a given distance bin since they may be located within a very similar distance from multiple other observation locations.
As a result, the estimation of confidence intervals for \(\widehat{err}_L^{(r)}(M)\) cannot be addressed with standard parametric techniques.
Similar issues are known from the estimation of confidence intervals for empirical semivariograms, where resampling procedures have therefore been proposed \citep{clark.allingham.2011.svgm.ci, olea.2011.svgm.ci}, which could be adapted to SPEPs.

In this study, \(r\) was chosen randomly within a desired range of separation distances, and \(N=50000\) and \(N=25000\) repetitions were used to obtain sufficient data for the estimation of profile functions in the Meuse and Maipo case studies, respectively.
A substantially smaller number may be sufficient and will be optimized in the future.
Binning was based on quadratically increasing breakpoints in order to show more detail at shorter distances.
The resulting error estimates were slightly smoothed with a weighted moving average.

\subsection{Spatial variable importance profiles (SVIPs)}\label{spatial-variable-importance-profiles-svips}

Model-agnostic tools that aid in the interpretation of ML models are a key topic in explainable artificial intelligence research, and permutation-based techniques are a simple and popular approach to this end \citep{molnar.2019.iml.book}.
Permutation-based variable importance (PVI) is defined as the decrease in model accuracy (or increase in error) obtained when making predictions \(\tilde{y}_{(k)}\) using permuted or `scrambled' feature values.
Specifically, a model is first fitted using the undisturbed training data; its performance is measured on test data.
Then, a feature's data is replaced with a randomly permuted series of data values, and the model's performance \(\widetilde{err}_L^{(r)}(M)\) is measured a second time using this data, \(\{(\tilde{y}_{(k)}, y_{(k)}, d_{(k)}), k\in B\}\).
This is repeated multiple times for each predictor, and the mean decrease in predictive accuracy is measured for each variable \citep{molnar.2019.iml.book}.
This algorithm can be embedded in resampling-based model assessment procedures such as the bootstrap \citep{breiman.2001.randomforest} and CV, including spatial CV \citep{russ.brenning.2010.svi}.
It can also be applied to trained models that cannot be retrained, although with the disadvantage, in the present context, that the separation distance cannot be controlled.

One decision to make in permuting a predictor variable is where to take the candidate values from: (1) the test sample, (2) the training sample, or (3) the entire dataset.
In LOO estimation, the test sample contains only one observations, which is why the entire dataset is chosen as a source of resampled values.

Similar to the estimation of SPEPs, the SVI at a specific separation distance is obtained by first binning the distances, and then estimating the prediction errors \(\widehat{err}\) and \(\widetilde{err}\) within these bins from undisturbed and permuted data.
The SVIP is obtained from these performance differences.

Spatial LOO estimation with varying separation distances creates an opportunity to assess the contribution of each predictor in a spatially differentiated manner.
Specifically, \emph{spatial variable importance profiles} (\emph{SVIPs}) target the following questions:

\begin{itemize}
\item
  Which predictors contribute the most to a model's spatial prediction skill at a given prediction distance?
\item
  Which predictors continue to be informative at greater distances, when extrapolating from the study region into uncharted terrain?
\item
  How do models differ in their ability to exploit information related to predictors and/or location?
\end{itemize}

\subsection{Beyond geographic space}\label{beyond-geographic-space}

Distance concepts, and prediction models that try to overcome distance, are not only relevant in geographic space, but also in time, in space--time, in three-dimensional physical space, in phylogenetic trees, and in feature space, to name only a few such settings that are relevant to the spatial and environmental sciences.
These emerging fields of distance-based CV estimation have been reviewed at depth by \citet{roberts2017}, and therefore only selected aspects are highlighted here as they relate to a possible adoption of distance-based prediction error and variable importance profiles in these applications.

In space--time, it is important to recognize the conceptual and also mathematical differences between distances in geographic space and in time \citep{cressie.wikle.2011}.
As a consequence, resampling schemes must be designed with a specific prediction objective in mind, for example hindcasting, forecasting, or regionalization.
While recently proposed space--time resampling schemes focused on hindcasting and regionalization, or a combination of both \citep{meyer2018}, forecasting has received particular attention in time-series research, where the consideration of an appropriate forecasting horizon has been identified as a critical issue in assessing model performance \citep{bergmeir.benitez.2012}.
The relationship between predictibility and lead time has received much attention in climatology \citep{palmer.hagedorn.2006}.
Similar to the spatial tasks studied here, it is therefore of critical importance to choose suitable spatial, temporal, or spatio-temporal distance metrics and prediction horizons in the assessment of spatio-temporal ML models instead of estimating performances based on `target' locations or times that are dictated by the sample itself, as in LOO-CV.
In this context, the distinction between hindcasting and forecasting is of particular importance since the effects of events or external stimuli propagate into the future, not into the past.

In three-dimensional geological space, the vertical dimension often represents, to some extent, time (for example, in stratigraphic sequences), and in the atmospheric sciences, air masses are stratified mainly vertically and not horizontally.
The concept of prediction distance may therefore depend on the specific application setting.
In the case of geometric anisotropy, this can be accounted for by means of a linear coordinate transformation that defines a common distance metric across all three dimensions \citep{cressie.wikle.2011}.

Distance-based profiles can further be applied in feature space using an appropriate distance metric in individual or multiple predictors \citep[environmental blocking,][]{roberts2017}.
Given the dependencies among predictor variables, the Mahalanobis distance appears to be a reasonable choice in situations involving quantitative (real-valued) predictors.

Distance-based profile functions of prediction error and variable importance, as proposed here, may be a useful tool in many of these situations.
Meaningful concepts of prediction distance can be defined based on, for example, the forecasting horizon, the vertical or three-dimensional physical distance, or graph-theoretical distances in a phylogenetic tree \citep{roberts2017}, in a stream network \citep{skoien.et.al.2006}, or in a social network.
An extension of the proposed prediction error and variable importance profiles to these distance metrics is straightforward and will not be further explored in this study.

\subsection{Implementation}\label{implementation}

The proposed methods were implemented in the open-source data analysis environment R using the \emph{sperrorest} package, which provides a flexible framework for resampling-based model assessment \citep{brenning.2012.sperrorest}.
The code is available in a GitHub repository under an open-source licence (\url{https://github.com/alexanderbrenning/spdiag}) and will be integrated into \emph{sperrorest} to provide additional user-level functions for the estimation and visualization of SPEPs and SVIPs.

\section{Case Study 1: regionalization using ML and geostatistics}\label{case-study-1-regionalization-using-ml-and-geostatistics}

The first case study is a well-known dataset on topsoil heavy-metal concentration on a floodplain of the Meuse river in the Netherlands as included in the \texttt{sp} package in R \citep{pebesma.bivand.2005.sp}.
It is widely used to introduce geostatistical interpolation techniques and demonstrate the combination of kriging with regression.

The combined use of spatial predictor variables, often derived from digital terrain models or remotely-sensed data, and spatial autocorrelation to spatially predict (or `regionalize') a quantitative response variable, is a common task in environmental science.
This case study is a typical use case for kriging with external drift \citep{cressie.1993} as well as for models that only exploit the information available from the predictor variables.

This study explores spatial prediction skill and variable importances of a selection of regionalization techniques that is intended to cover a broad spectrum from pure interpolation to spatial and non-spatial ML.

\subsection{Case study description: the Meuse dataset}\label{case-study-description-the-meuse-dataset}

The Meuse dataset contains 155 observations of (logarithmic) topsoil zinc concentration (\emph{logZn} in log-ppm) as the response variable, and several possible predictor variables.
Zinc concentrations in this study area are related to the amount of contaminated sediment deposited on the floodplain, and therefore to predictors such as elevation (\emph{elev}) and distance to river.
This study uses these predictor variables, applying a square-root transformation to distance (\emph{sqrt.dist}), in addition to UTM \emph{x} and \emph{y} as predictors that represent possible spatial trends.
A linear model with these four predictors explains 72.6\% of the variance of \emph{logZn}, and has a residual autocorrelation range of 926~m with a nugget-to-sill ratio of 0.27.
For comparison, \emph{logZn} itself has a range of 897~m with a nugget-to-sill ratio of 0.27, making it also very suitable for ordinary kriging interpolation without any trend predictors.

The floodplain is approximately 4 km \(\times\) 1 km in size.
The average nearest-neighbour distance of sampling locations is 112~m (minimum: 44~m).
If the goal is to make spatial predictions on the floodplain itself, which is the usual use case for this dataset in the literature, it should be noted that the average prediction distance is 96~m (1st, 3rd quartiles: 53 and 120~m; see Appendix~\ref{sec:preddist}).
Mean nearest-neighbour distance and mean prediction distance are similar in this case study due to the relatively uniform, nearly random distribution of sampling locations on the floodplain.

\subsection{Regionalization techniques and their assessment}\label{cs1assessment}

In this case study, spatial diagnostics of the following contrasting spatial prediction methods were compared:

\begin{enumerate}
\def\labelenumi{\arabic{enumi}.}
\item
  Nearest-neigbour interpolation (NN) was chosen as a simple deterministic baseline method.
\item
  Ordinary kriging (OK) was included as a basic geostatistical technique without predictors of trend \citep{cressie.1993}.
\item
  Kriging with external drift (KED, or universal kriging), is a geostatistical technique that incorporates all four variables as linear predictors \citep{cressie.1993}.
\item
  Multiple linear regression (MLR) using the same four predictors was included as it (also) models a linear trend, but it does not exploit spatial dependence.
\item
  Geographically weighted regression (GWR) was selected as a locally linear model with spatially varying coefficients \citep{fotheringham.et.al.2002}.
\item
  Random forest (RF) was chosen as it is a popular nonlinear ML technique that is agnostic of the spatial setting \citep{breiman.2001.randomforest}.
\item
  A combined OK--RF model was furthermore designed as an experimental hybrid geostatistical--ML technique that fades linearly from a pure OK interpolation at 0~m prediction distance to a pure RF at \(\ge 500\)~m distance.
\end{enumerate}

\begin{figure}
\includegraphics[width=1\linewidth]{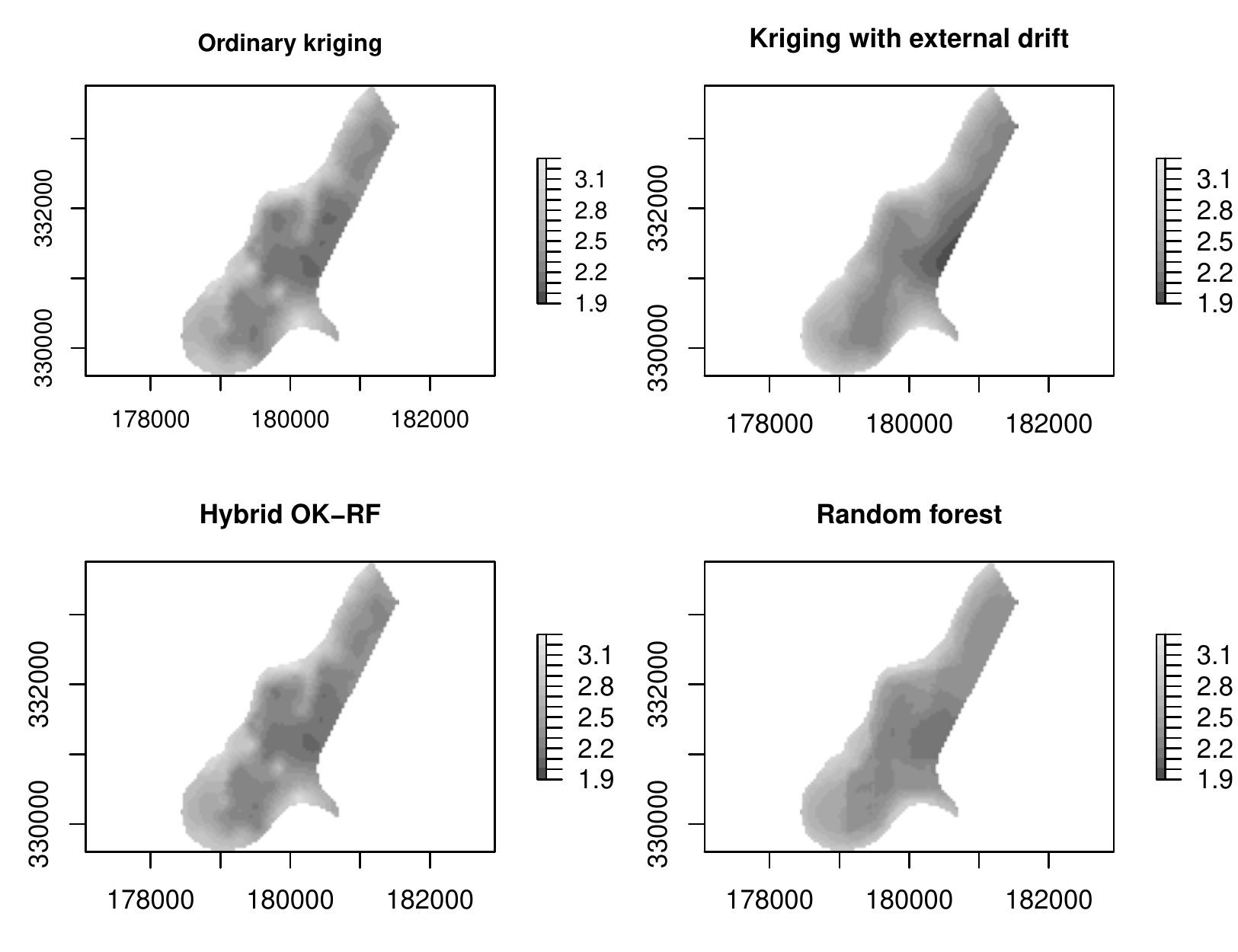} \caption{Spatial prediction maps of $logZn$ in the Meuse case study using four selected geostatistical, ML, and hybrid models.}\label{fig:meuseprediction}
\end{figure}

Model parameter values and implementation details are reported in Appendix~\ref{sec:modeldetails}. The chosen algorithm settings were not optimized since the present study is not a benchmarking exercise.
The models were rather selected to illustrate the spatial behaviour of various model types.
In particular, OK--RF was built for pedagogical reasons as the proposed SPEPs and SVIPs should be able to highlight the constrasting short- and long-distance behaviours of this model.
Figure \ref{fig:meuseprediction} shows the prediction maps obtained with OK, KED, OK--RF, and RF.

In order to contextualize SPEPs in the broader context of spatial model assessment, various other error estimators were also calculated:

\begin{enumerate}
\def\labelenumi{\arabic{enumi}.}
\item
  Resubstitution error, estimated on the training sample, which is inherently overoptimistic;
\item
  LOO-CV at the level of sample locations, ignoring spatial autocorrelation;
\item
  Non-spatial, random 10-fold CV at the level of point observations, with the same limitation;
\item
  Spatial 10-fold CV using 10-means clustering to partition the study region \citep{russ.brenning.2010}.
  Both types of 10-fold CV were repeated 50 times.
\end{enumerate}

\subsection{Spatial prediction error profiles}\label{spatial-prediction-error-profiles}

\begin{figure}
\includegraphics[width=1\linewidth]{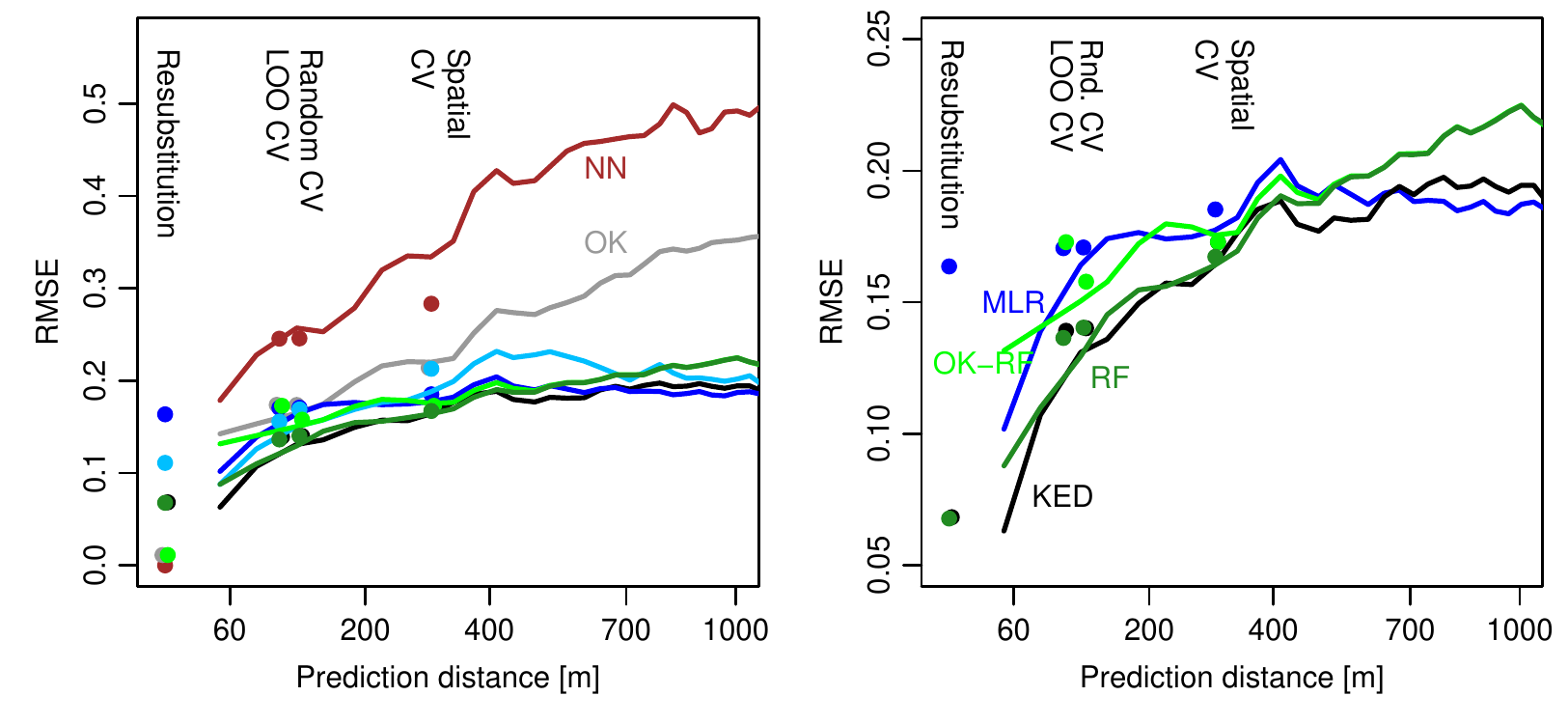} \caption{Spatial prediction error profiles for the prediction of logZn in the Meuse case study. The x axes are square-root transformed. LOO-CV and other point estimators are placed close to their mean prediction distance. Left: All models; right: detailed view of KED, MLR, RF, and OK--RF. Models: NN (brown), OK (grey), KED (black), MLR (blue), GWR (light blue), RF (dark green), OK--RF (light green).}\label{fig:meuseerrorprofile}
\end{figure}

In the Meuse case study, the SPEPs revealed a strong dependence of performance on prediction distance for all methods, with some surprising similarities between (geo-)statistical and ML techniques (Figure \ref{fig:meuseerrorprofile}).

Overall, interpolation techniques that do not incorporate predictor variables (NN, OK), had higher prediction errors especially at greater prediction distances.
In general terms, OK's increase in RMSE with distance is consistent with the skill expected based on the target variable's semivariogram (square root of nugget effect: 0.10, of total sill: 0.35), which provides a rough indication of the kriging prediction error at near-zero and at long prediction distances (greater than the autocorrelation range).

At short prediction distances (up to about 300~m), KED, GWR, and RF displayed relatively similar error profiles, closely followed by MLR.
This is unexpected considering their completely disparate approaches to spatial prediction.
Considering the distribution of prediction distances on the floodplain from the sampling locations, KED had an edge over RF, which was followed by GWR.

At greater distances, GWR and RF (and OK--RF, which is identical to RF at distances \(\ge 500\)~m) showed substantially higher prediction errors, indicating a poorer spatial transferability compared to the simpler linear models underlying KED and MLR.

The weakest, but still noticeable dependence of performance on distance was found in MLR.
The relatively large resubstitution error of MLR also underlines the fact that the limited flexibility limits its ability to fit ---or to overfit--- to local patterns or anomalies.
Note that the drop in RMSE towards small separation distances is not in contradiction to this.
If a left-out observation represents, for instance, a positive anomaly, then nearby observations will also tend to have above-average values.
This will pull the prediction surface ---for example, through a change in MLR's intercept--- towards larger predicted values, which will thus reduce the RMSE at short distances.

This indirect effect of spatial autocorrelation on non-spatial MLR and RF models can therefore explain distance-dependent variation in predictive performances of non-spatial models.
It was more pronounced in RF than in MLR since RF was better able to (over-)fit to to the training sample ---a behaviour that appears to be beneficial in short-distance regionalization in the presence of strong spatial autocorrelation.

Figure \ref{fig:meuseerrorprofile} also highlights limitations of the proposed approach. At distances shorter than the minimum nearest-neighbour distance (44~m), it will inevitably have a blind spot.
But even at slightly greater separation distances around the median nearest-neighbour distance (107~m), only limited and possibly geographically biased data may be available for estimating spatial prediction skill.
In this study, below-average nearest-neighbour distances occur throughout the study area, with the exception of the southeasternmost fringe.

Considering the longer prediction distances, in this case study, robust training sample sizes were available for all displayed distances. The average training sample size (out of \(n=155\)) dropped below 140 for separation distances \textgreater415~m, and below 100 only for \textgreater1000~m distance in spatial LOO resampling.

\subsection{Comparison to other performance estimators}\label{comparison-to-other-performance-estimators}

As expected, spatial CV estimates of model performance showed a consistently larger prediction error than non-spatial CV estimates.
This can be attributed to the greater mean separation distance between test and training locations of 298~m for spatial CV versus 116 and 112~m for non-spatial CV and LOO-CV, respectively.
Considering the relatively uniform sample distribution, surprisingly the mean prediction distance throughout the entire floodplain (i.e., at unsampled locations) of 96~m was even smaller than the mean prediction distances of CV estimators.
These differences in mean separation distances, which are usually not reported along with CV estimates, underline the need to assess model performance at specific distances more explicitly, and in a more targeted manner.

When comparing the various models, spatial and non-spatial CV estimators alike placed RF among the top 1-2 models, along with KED.
However, due to their implicit focus on specific separation distances, these estimators failed to detect the consistently larger RMSE of RF at distances \textless100~m (+122\% compared to KED) and \textgreater500~m (about +110\% compared to KED and MLR). Although the practical relevance of the increase at large versus short separation distances will depend on the application scenario at hand, overall only the SPEPs revealed the outperformance of RF by KED in this case study.

\subsection{Spatial variable importance profiles}\label{spatial-variable-importance-profiles}

\begin{figure}
\includegraphics[width=1\linewidth]{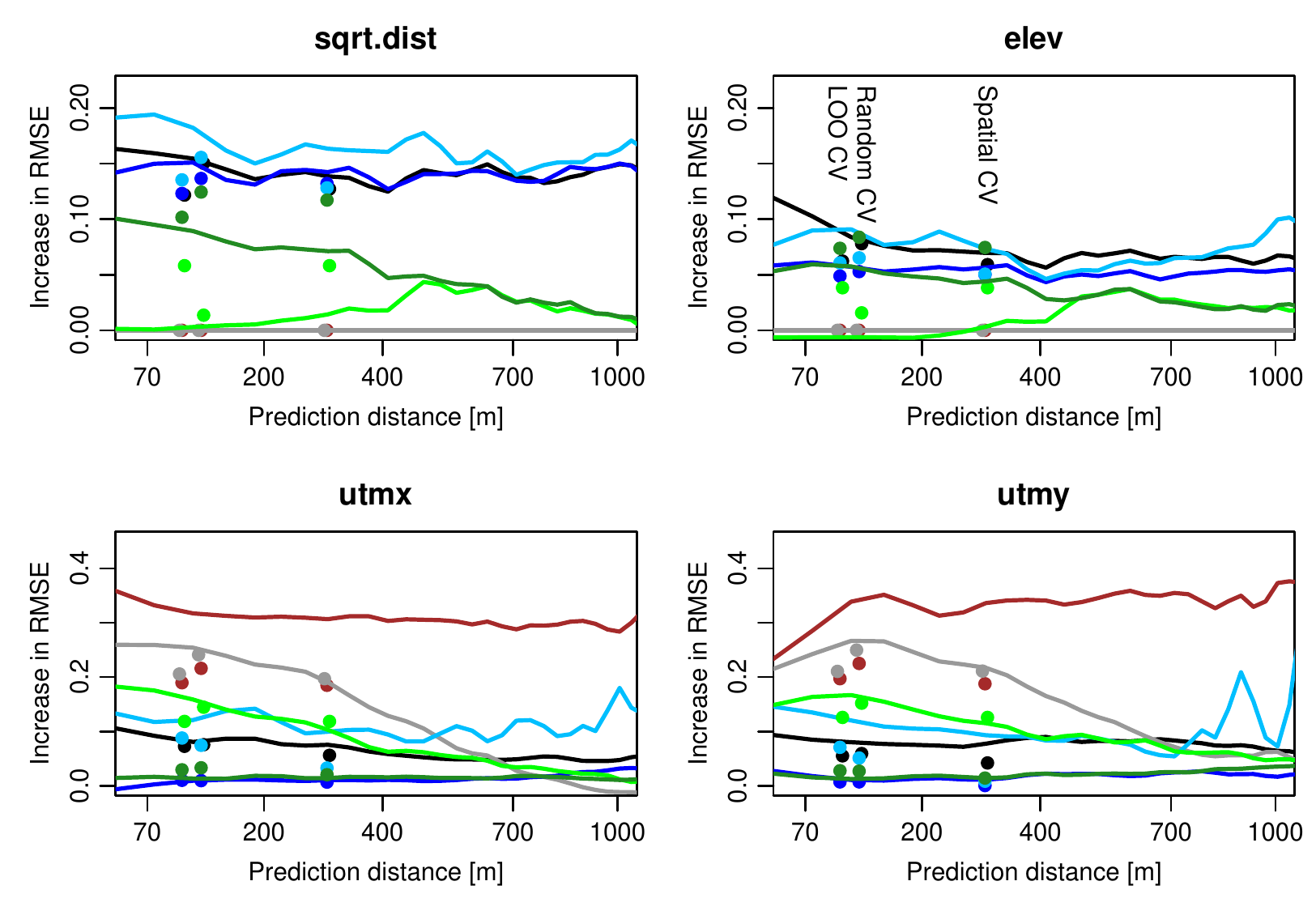} \caption{Spatial variable importance profiles in the prediction of $logZn$ in the Meuse case study. Models: NN (brown), OK (grey), KED (black), MLR (blue), GWR (light blue), RF (dark green), OK--RF (light green).}\label{fig:meusesviprofile}
\end{figure}

Interpolation, regression, and hybrid models displayed a clear difference in the relative importance of \emph{sqrt.dist} and \emph{elev} compared to the \emph{x}/\emph{y} coordinates.
Specifically, MLR, KED and GWR showed, on average, very similar, distance-insensitive importance profiles for \emph{sqrt.dist} and \emph{elev}.
OK predictions were more dependent on location for short prediction distances than at greater ones, where spatial autocorrelation weakens and OK predictions approach the overall sample mean \citep{cressie.1993}.
The NN method, in contrast, mathematically does not exhibit this averaging behaviour, and consequently the empirically estimated importance of \emph{x} and \emph{y} remained independent of distance.
OK's and NN's inability to account for spatial trends is documented by flat SVIP lines on zero, which explains the poor predictive skill at long distances.

SVIPs of RF are furthermore instructive as they help to explain its distance-dependent prediction error.
RF's importance of \emph{sqrt.dist} and \emph{elev} increased continuously towards shorter prediction distances, which is quite remarkable considering the non-spatial design of standard RF.
As discussed in the previous section, it appears that this is evidence of RF's excellent ability to learn, and perhaps memorize, small variations in regression relationships, which are only valid locally near the training locations.
This explains how RF implicitly benefits from spatial autocorrelation, even without explicitly exploiting prediction distance information.

OK--RF, in constrast, was deliberately designed to mainly rely on spatial autocorrelation at short distances via the use of OK, and to fade into a RF model up to a distance of 500 m.
The SVIPs reflect this model structure very clearly as the importance of \emph{sqrt.dist} and \emph{elev} ramps up from 0 to 500 m prediction distance, where they become identical to the SVIP curves of RF.
Similarly, OK--RF's SVIPs of \emph{x} and \emph{y} show a decreasing trend.
They do not reach exactly zero; this is a consequence of the permutation method randomly assigning permuted \emph{x} and \emph{y} values that may correspond to short prediction distances, which in turn switches the OK--RF model from RF to OK mode.

\subsection{Comparison to other importance estimators}\label{comparison-to-other-importance-estimators}

Again, due to the relatively short mean prediction distances of established spatial and non-spatial resampling techniques, only the proposed SVIPs were able to make relevant distance-related differences in variable importance visible.
In the case of OK--RF, however, these variable importance estimates were surprisingly inconsistent with the SVIP-based assessment, which were the most reliable diagnostics for detecting the (in this case, known) spatial structure of this experimentally designed hybrid model.

\section{Case study 2: spatial classification}\label{case-study-2-spatial-classification}

Crop classification using multispectral satellite image time series is a broad and important ML task in environmental remote sensing.
Knowledge of SPEPs is important in order to assess the potential of classifiers to be applied in adjacent study regions.
This involves a large number of correlated predictors representing vegetation phenology, which are difficult to analyze separately but can be projected into a lower-dimensional transformed space for better visualization \citep{brenning.2021.wiml}.

\subsection{Case study description: the Maipo dataset}\label{case-study-description-the-maipo-dataset}

The dataset used is a well-documented case study consisting of 400 fields (7713 grid cells in total) with 4 different fruit-tree crops in central Chile \citep{pena.brenning.2015.maipo}.
To simulate use cases with typical learning sample sizes, data from 100 fields (25 from each crop type) was sampled repeatedly, and results were averaged.
The feature set comprises 48 features representing visible and near- to shortwave-infrared spectral reflectances from Landsat images taken at 8 time points during one growing season, and 16 derived spectral indices.
Specifically, the normalized differences vegetation index (NDVI) and the normalized differences water index (NDWI) were included. Refer to \citet{pena.brenning.2015.maipo} for details.

These features are strongly correlated with each other, especially for subsequent time points (since fruit-tree characteristics don't change dramatically within a few weeks), and physiologically or mathematically related features. Correlation is particularly strong among image dates 1 and 2 (early-season features), within image date 3 (mid-season), and among dates 4-8 (late season).

\subsection{Classifiers and their assessment}\label{classifiers-and-their-assessment}

In this case study, spatial diagnostics of three contrasting spatial and non-spatial classifiers were compared:

\begin{enumerate}
\def\labelenumi{\arabic{enumi}.}
\item
  Random forest (RF) was chosen since it is a popular nonlinear technique that is widely used in remote sensing \citep{breiman.2001.randomforest, pal.2005}.
\item
  Linear discriminant analysis (LDA) was included as it is a simple but robust classification technique.
\item
  A combination of (spatial) nearest-neighbour classification (at prediction distances \(\le 100\)~m) with LDA (at greater distances) is included as a simple, illustrative approach that uses only spatial proximity or only remotely-sensed features, depending on target distance (NN--LDA).
\end{enumerate}

Only the third, illustrative technique is designed to explicitly account for spatial dependence in the data. Details are given in the Appendix.

SPEPs and SVIPs were calculated for separation distances ranging from 30 m (i.e., grid resolution) to over 10 km (diameter of study area: about 40 km) using the misclassification rate as the error measure.

Similar to the Meuse case study, model performances and variable importances were furthermore estimated using other resampling-based techniques for comparison. In addition to LOO-CV, non-spatial CV and \(k\)-means-based spatial CV (see section \ref{cs1assessment}), a second type of spatial CV was used in which fields (i.e.~groups of grid cells) are resampled (field-level CV, the method used by \citep{pena.brenning.2015.maipo}).

Given the high dimensionality of the feature space and the strong correlations among features, the approach proposed by \citet{brenning.2021.wiml} was adopted to estimate variable importances from a transformed perspective.
Specifially, principal-component (PC) transformations were applied to feature subspaces spanned by early-, mid- and late-season predictors, respectively.
For convenience, only the SVIPs of the first PCs are presented.
SVI assessment in transformed PC space bypasses the problem that permutation techniques should not be applied to strongly dependent features \citep{hooker.mentch.2019.stop, brenning.2021.wiml}.

In the interpretation of the following results it is important to remember that predictions at less than 100~m distance and up to about 500~m distance may occur within fields that are already included in the training sample.
This could be relevant in gap-filling applications, but not in the usual setting of land cover classification in which `new' fields in the same or an adjacent region are to be classified.
Note that in this case study, the training sample size in the LOO procedure does not decrease with increasing separation distance since the same number of fields is always sampled from the large pool of data in the remaining area.

\subsection{Spatial prediction error profiles}\label{spatial-prediction-error-profiles-1}

\begin{figure}
\includegraphics[width=0.8\linewidth]{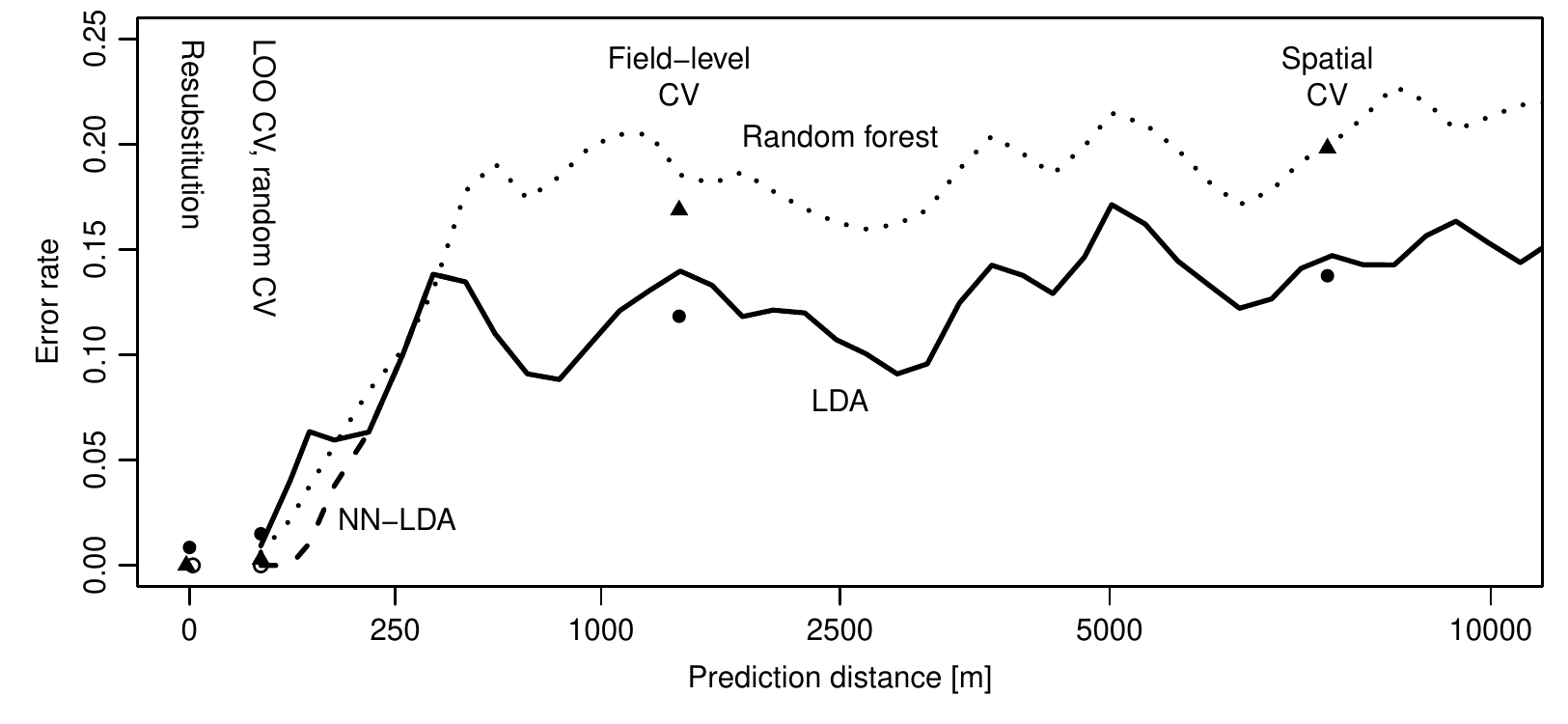} \caption{Spatial prediction error profiles for the classification of crop type in the Maipo case study (RF: dotted line, triangle; LDA: solid line, bullet; NN--LDA: dashed, empty circle). Point estimates from different CV types are plotted at their mean prediction distances. Results of LOO-CV and non-spatial random CV are visually indistinguishable.}\label{fig:maipoerrorprofile}
\end{figure}

The SPEPs of LDA and NN--LDA clearly highlight the capability of the proposed approach to detect spatial differences in prediction skill (Figure \ref{fig:maipoerrorprofile}).
By construction, NN--LDA will flawlessly classify crops up to 100~m distance whenever the target location lies within a field from the training sample.
At greater distances, NN--LDA is indistinguishable from LDA by design.
This behaviour was effectively detected with the help of the SPEP.

RF interestingly also showed a decrease in error rate towards the shortest prediction distances.
This can be attributed to the overfitting of RF to the training sample, which proves to be an advantage over LDA in within-field classification.
Nevertheless, RF did not achieve the same skill in within-field classification as NN--LDA, which leverages expert knowledge on the spatial structure of the classification task.
At distances beyond the field scale, LDA (and NN--LDA) outperformed RF, as detected previously with field-level CV \citep{pena.brenning.2015.maipo}.

\subsection{Comparison to other performance estimators}\label{comparison-to-other-performance-estimators-1}

The various CV estimators also showed important differences.
Overall, error rates estimated at short prediction distances, and non-spatial random CV in particular, grossly underestimated the regional-scale prediction error.
This holds true especially in the case of RF (due to overfitting) and NN--LDA (due to its spatial design).
Beyond the field scale, i.e.~at distances greater than about 500~m, error rates increased only slightly, which can be attributed to largely homogeneous agricultural and environmental conditions within this study region.

Considering the use case of classifying crops in the entire study region, prediction distances obtained with field-level CV resampling were most similar to regional-scale prediction distances (mean prediction distance of 871~m in field-level CV versus 831~m overall; see Appendix~\ref{sec:preddist} for histograms).
\(k\)-means-based spatial-CV resampling represented, on average, much larger prediction distances (6664~m).
These were consistent with spatial-LOO prediction errors at similar distances, but they come at a substantially lower computational cost than spatial LOO estimation, which requires fitting a new model for each LOO prediction.
Nevertheless, the results underline the importance of reporting mean prediction distances along with resampling-based performance measures, comparing them to the prediction distances of the application setting.

\subsection{Spatial variable importance profiles}\label{spatial-variable-importance-profiles-1}

\begin{figure}
\includegraphics[width=1\linewidth]{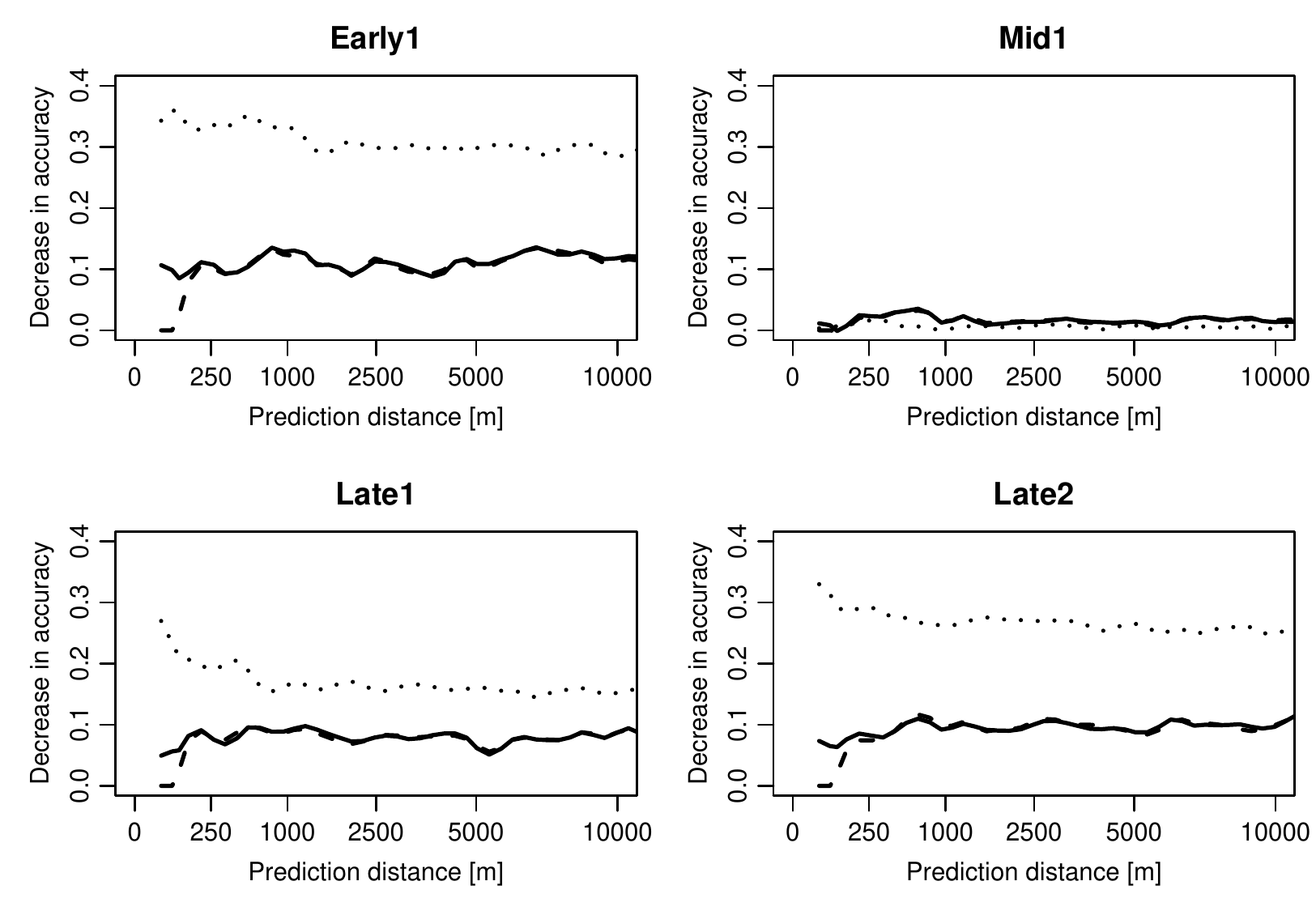} \caption{Spatial variable importance profiles in crop classification in the Maipo case study for the first PC of early- and mid-season features and the first two PCs of late-season features. Models: RF (dotted line), LDA (solid line), NN--LDA (dashed).}\label{fig:maiposviprofile}
\end{figure}

SVIPs were effectively able to detect the striking difference in model structure between LDA and NN--LDA (Figure~\ref{fig:maiposviprofile}).
They clearly indicated that at short distances, NN--LDA did not make use of the available predictors.
LDA and RF both showed increases in SVI at short distances for some groups of variables (e.g., first PC of late-season predictors, \emph{Late1}), which may be indicative of overfitting.
The generally higher SVIs in RF than in LDA (despite the poorer overall error rate) are attributed to a stronger concentration of RF in predictors associated with \emph{Early1}, \emph{Late1} and \emph{Late2} PCs, which only represent a fraction of the overall variance of the 64 available predictors.

\section{Discussion}\label{discussion}

\subsection{Distance-based spatial model assessment and interpretation}\label{distance-based-spatial-model-assessment-and-interpretation}

In the model-agnostic spatial model assessment, SPEPs demonstrated their ability to highlight strengths and weaknesses of different models in predicting the response locally, and in transferring the modeled relationships to more distant regions.
In combination with knowledge of the intended distribution of prediction distances, they allow modelers to make better-informed choices regarding model design and selection.
In spatial model assessments, we need to shift the focus from the question `At what distance does the test data become independent?' to `At what distances would I like to predict the response?'

With regards to the case studies, KED was superior at very short distances due to its mathematical optimality as a best linear unbiased predictor, but RF achieved very similar performances even at relatively short prediction distances despite its non-geostatistical nature.
In classification, SPEPs identified very sharply the contrasting behaviour of LDA and an experimentally designed hybrid NN--LDA that incorporates nearest-neighbour interpolation at short distances.
Again, RF showed a remarkable drop in prediction error towards short distances, which was interpreted as a positive side-effect of overfitting.

Given the ability of SPEPs to measure and visualize model performance seamlessly across scales, they bridge the scale gaps between resubstitution error, non-spatial model assessments, and different types of spatial assessment (e.g., field-level or region-based).
This reminds us that predictive performance strongly depends on the objective of the prediction, e.g.~local gap filling, or regional generalization and model transfer.
SPEPs have the potential to offer a more differentiated perspective on model performance than simple non-spatial assessments of models such as kriging, spatial regression, and random forests, which may exhibit contrasting spatial behaviour by design.
This has not been acknowledged sufficiently in previous benchmarking studies that (1) either ignore the scale of spatial prediction \citep{goovaerts.2000.interpolation, fox.et.al.2020}, (2) postulate that spatial dependence needs to be taken care of based on the range of residual autocorrelation \citep{brenning2005, roberts2017, valavi2019}, or (3) focus on other fixed scales of spatial prediction \citep[e.g.,][]{pena.brenning.2015.maipo, goetz.et.al.2015.cg}.
After all, the spatial scale of model assessments should not be dictated by some poorly defined (residual?) spatial autocorrelation criterion, but by the purpose of the prediction.

In the context of model interpretation, SVIPs proved to be capable of identifying spatially aware model structures as present in the combined NN--LDA and OK--RF techniques examined in classification and regionalization, respectively.
Despite the well-known limitations of permutation-based variable importance measures, this novel approach offers a simple and intuitive approach to spatially differentiated model-agnostic interpretation of ML models.
It may also serve as a template for spatially nuancing other diagnostic tools for explainable ML, and for developing similar approaches in the spatio-temporal domain or in feature space.

\subsection{Computational versus theoretically motivated measures of spatial model performance}\label{computational-versus-theoretically-motivated-measures-of-spatial-model-performance}

Theoretically derived measures of uncertainty such as kriging variances or prediction intervals of linear regression models provide a reliable uncertainty assessment when their model assumptions are satisfied.
In the regionalization case study, computational SPEPs were consistent with theoretical expectations where available (OK, KED).
Geostatistically based performance measures have a low computational cost, but it may be difficult to judge the effects of the possible violation of model assumptions.
As a matter of fact, OK's stationarity assumption is violated when a trend or external drift is present, as in the Meuse case study, and therefore kriging variances output by OK and KED cannot be compared directly.
Computational tools as presented in this study, in contrast, are model-agnostic and therefore allow us to compare different algorithms independently of their underlying assumptions and paradigms, and regardless of whether or not we believe they are satisfied.
They offer a data-driven second opinion on prediction performance even where model-based variance estimates are available.

Nevertheless, we should acknowledge that SPEPs that only depend on distance, as presented here, only provide a simplified, stationary and isotropic perspective on predictive model behaviour, as only distance, and not location or orientation, is taken into account.
Unlike the semivariogram in geostatistics, which can be estimated for specific directions by filtering suitable pairs of points, the directional approach cannot be transferred to the estimation of SPEPs and SVIPs.

\subsection{The role of autocorrelation and independence in spatial model assessment}\label{discussionscale}

It has previously been proposed to choose the buffer distance based on the range of residual autocorrelation \citep{brenning2005, rest2014, valavi2019}.
Nevertheless, this starts from the intuition that test samples must be independent, although often without providing a precise definition \citep[e.g.,][]{pohjankukka2017, valavi2019}.
What makes things worse is that the range of autocorrelation of the residuals will inevitably be model-dependent, and in the case of overfitting models, they provide a biased estimate of model error and consequently of the autocorrelation range.

From a practical perspective, independent test data is not even a desirable property in predictive tasks such as interpolation or regionalization that precisely build upon and require spatial dependence \citep{cressie.1993}.
Not only geostatistical models but also hybrid ML interpolation techniques increasingly exploit this dependence \citep[this study and][]{sekulic.et.al.2020}.
In these situations, spatial prediction uncertainty will inevitably depend on prediction distance or horizon, which we must therefore incorporate in our model assessment and interpretation, as proposed in this study.

\section{Conclusions}\label{conclusions}

The proposed distance-based spatial model assessment and interpretation tools enrich the toolkit available for explaining the decisions of ML models in the spatial domain.
They produce intuitively interpretable visualizations of the spatial transferability of modeled relationships.
Results obtained in two environmental-science and remote-sensing case studies were encouraging and identified important differences as well as similarities among various statistical, geostatistical, ML, and hybrid models.
SPEPs and SVIPs were effectively able to identify key differences, in particular whether a black-box model exploits proximity information or entirely relies on predictor--response relationships.

Compared with increasingly popular resampling-based spatial model assessments with a fixed spatial block size, the continuous-distance-based approach offers substantially more detail and relates performance directly to prediction distance.
In this context, the practice of using the range of residual spatial autocorrelation as a minimum separation distance should be abandoned as it lacks a coherent theoretical justification and is not derived from the spatial prediction task at hand, which may specifically exploit spatial dependence.
The distance-dependence of performance further implies that mean prediction distances of prediction tasks and of performance estimators such as CV should be comparable. They should routinely be reported in spatial prediction modeling.

It is suggested that the wider use of spatially aware model-assessment and interpretation tools has the potential to improve the practice of spatial prediction modeling in fields ranging from remote sensing to ecology and the environmental sciences.
Data-driven diagnostics provide a valuable, assumption-free second opinion on model performance even in situations where theory-based prediction variances are available.
These tools furthermore generate opportunities for designing improved classification and regionalization models that focus on a clearly defined spatial prediction horizon.


\pagebreak

\appendix

\section{Distribution of prediction distances}\label{sec:preddist}

\begin{figure}[h]
\includegraphics[width=1\linewidth]{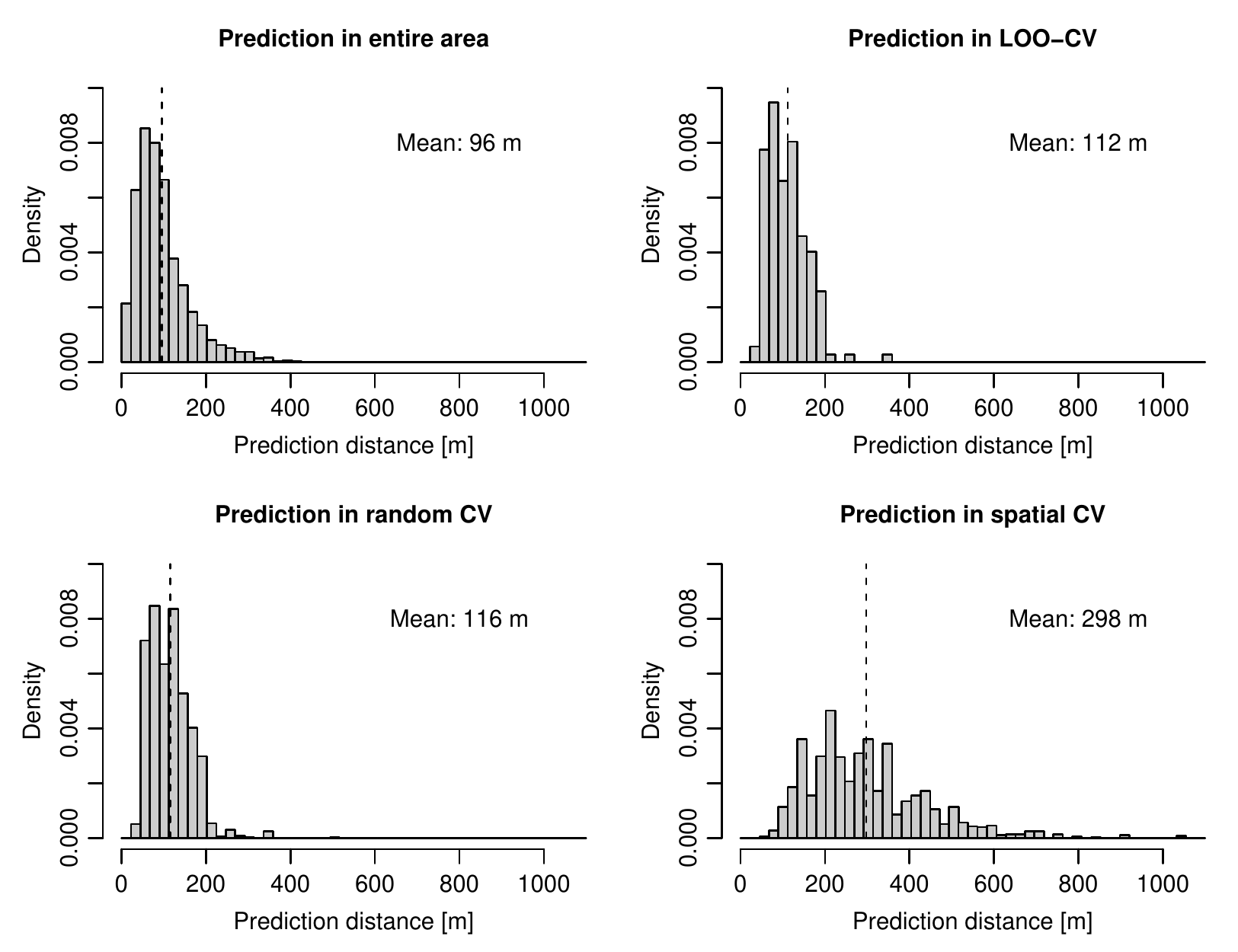} \caption{Histograms of prediction distances in the Meuse case study. Top left: Distribution based on the 155 training locations and 3103 target grid cells as prediction locations. Top right: LOO-CV. Bottom left: Non-spatial (random) CV. Bottom right: $k$-means-based spatial CV. Distributions for CV-based estimators are based on 50 repetitions.}\label{fig:meusepreddist}
\end{figure}

\begin{figure}[h]
\includegraphics[width=1\linewidth]{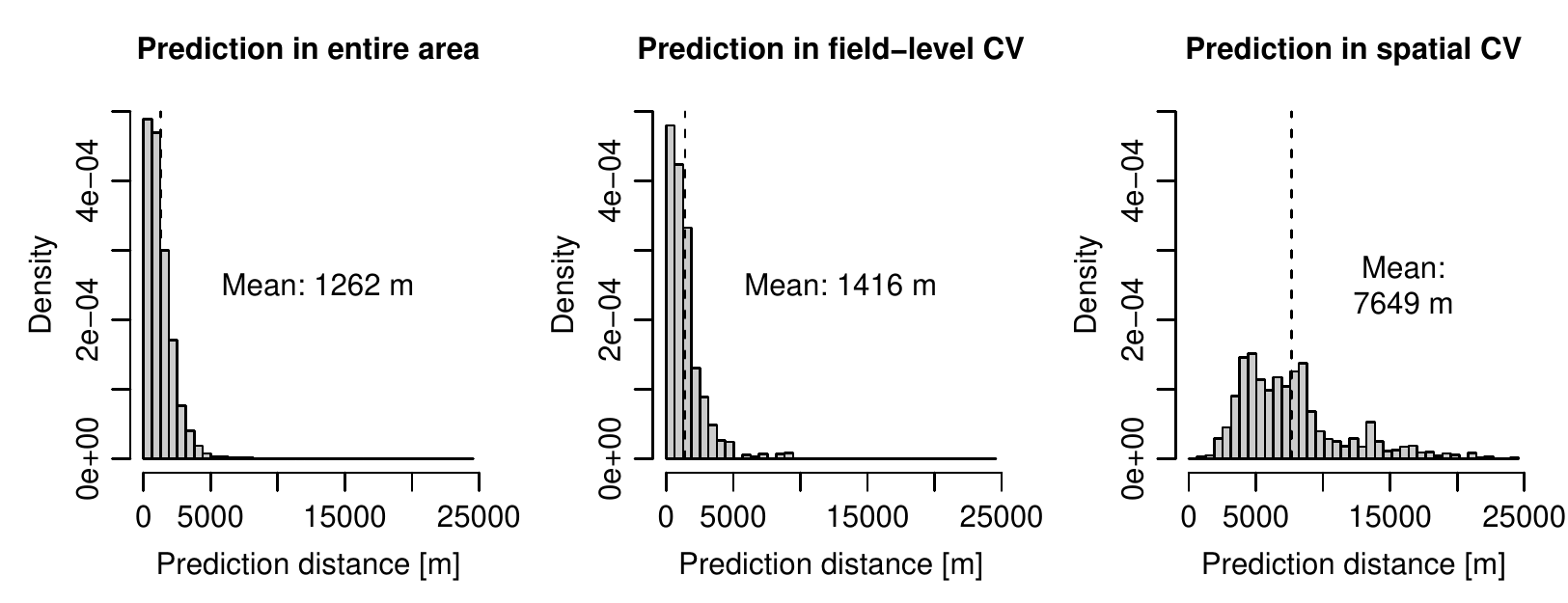} \caption{Histograms of prediction distances in the Maipo case study. Left: Distribution based on randomly selecting 100 fields for training and the remaining 300 fields as prediction locations. Center and right: Distributions for field-level CV and $k$-means-based spatial CV, respectively. All histograms are aggregated over 50 samples or CV repetitions.}\label{fig:maipopreddist}
\end{figure}

The histograms in Figures \ref{fig:meusepreddist} and \ref{fig:maipopreddist} display the distributions of prediction distances in both case studies in different validation scenarios and.
For comparison, the diagrams show the distribution of prediction distances in the prediction task in which a ML model trained on the/a training sample is used to predict the response variable throughout the entire study area.

\section{Model details}\label{sec:modeldetails}

NN interpolation was implemented based on the observed value of the (one) nearest observation according to the Euclidean distance in UTM coordinate space.

OK was applied as a global interpolation technique with a spherical semivariogram model that was re-fitted to each training sample using iteratively re-weighted least squares \citep{cressie.1993}.
Empirical semivariograms were estimated each time with Cressie's robust estimator.
KED's residual semivariogram was similarly modeled with a spherical model fitted to a robust empirical residual semivariogram.
Microscale variability was represented as a measurement error variance instead of the more common nugget effect.
This turns OK and KED into smoothers, while a nugget effect would produce a local discontinuity at the training locations.
This only affects the reported resubstitution errors, which would be 0 otherwise.
The OK and KED implementation in R's \emph{gstat} package were used \citep{graeler.et.al.2016.gstat}.

GWR in the R implementation in package \emph{spgwr} \citep{bivand.yu.2020.spgwr} was used with an inner CV for optimizing the bandwidth parameter.
A global bandwidth parameter instead of local ones was chosen to reduce the probability of overfitting.
Only \emph{sqrt.dist} and \emph{elev} served as predictors in this model; UTM \emph{x} and \emph{y} were only used to define the geographic space within which the coefficients vary.

The RF model from the original R implementation in the \emph{randomForest} package was used with default settings, such as 500 trees \citep{liaw.wiener.2002.randomforest}.

OK--RF regionalization was implemented by fitting OK and RF models as described above, and by combining their predictions \(\hat{y}_{OK}\) and \(\hat{y}_{RF}\) depending on the prediction distance \(d\), up to a maximum distance \(d_{\textrm{max}}:=500\)~m:
\[
\hat{y}_{\textit{OK-RF}} := \rho\hat{y}_{RF} + (1-\rho)\hat{y}_{OK},
\]
where \(\rho = \textrm{min}\{d/d_{\textrm{max}}, 1\}\).
No attempt was made to optimize \(d_{\textrm{max}}\) as this model was primarily designed for demonstration purposes.

LDA classification was based on the implementation in the \emph{MASS} package \citep{venables.ripley.2002}.

NN--LDA was implemented based on the LDA classifier and a (one-)nearest-neighbour classifier that uses Euclidean distance in UTM coordinate space to measure proximity.
The classifier switches from nearest-neighbour to LDA mode at a separation distance of 100~m with no transition zone.
This simple and transparent setting was chosen for illustrative purposes.
No attempt was made to optimize the threshold distance, which is nevertheless generally consistent with typical field sizes in the study area, or to implement a transition zone.

\end{document}